\def\year{2021}\relax
\documentclass[letterpaper]{article} 
\usepackage{aaai21}  
\usepackage{times}  
\usepackage{helvet} 
\usepackage{courier}  
\usepackage[hyphens]{url}  
\usepackage{graphicx} 
\usepackage{natbib}  
\usepackage{caption} 
\nocopyright
\usepackage{multirow}
\usepackage{xcolor}
\usepackage{amsmath}
\usepackage{algorithm}
\usepackage{algpseudocode}
\usepackage{blindtext}

\usepackage[capitalise]{cleveref}

\usepackage{amsmath,amsfonts,bm}

\def\eqref#1{equation~\ref{#1}}

\def\1{\bm{1}}

\def\vx{{\bm{x}}}
\def\vy{{\bm{y}}}

\DeclareMathAlphabet{\mathsfit}{\encodingdefault}{\sfdefault}{m}{sl}
\SetMathAlphabet{\mathsfit}{bold}{\encodingdefault}{\sfdefault}{bx}{n}
\newcommand{\tens}[1]{\bm{\mathsfit{#1}}}
\def\tA{{\tens{A}}}

\def\tD{{\tens{D}}}

\def\tM{{\tens{M}}}

\def\tW{{\tens{W}}}

\newcommand{\R}{\mathbb{R}}

\title{KSM: Fast Multiple Task Adaption via Kernel-wise Soft Mask Learning}
\author{Li Yang, Zhezhi He, Junshan Zhang, Deliang Fan\\} 
\affiliations{School of Electrical, Computer and Energy Engineering, Arizona State University, Tempe, AZ\\
\{lyang166, zhezhihe, Junshan.Zhang, dfan\}@asu.edu}

\begin{document}
\maketitle

\begin{abstract}
Deep Neural Networks (DNN) could forget the knowledge about earlier tasks when learning new tasks, and this is known as \textit{catastrophic forgetting}. 
While recent continual learning methods are capable of alleviating the catastrophic problem on toy-sized datasets, some issues still remain to be tackled when applying them in real-world problems. Recently, the fast mask-based learning method (e.g. piggyback \cite{mallya2018piggyback}) is proposed to address these issues by learning only a binary element-wise mask in a fast manner, while keeping the backbone model fixed. However, the binary mask has limited modeling capacity for new tasks. A more recent work \cite{hung2019compacting} proposes a compress-grow-based method (CPG) to achieve better accuracy for new tasks by partially training backbone model, but with order-higher training cost, which makes it infeasible to be deployed into popular state-of-the-art edge-/mobile-learning. 
The primary goal of this work is to simultaneously achieve fast and high-accuracy multi task adaption in continual learning setting. Thus motivated, we propose a new training method called \textit{kernel-wise Soft Mask} (KSM), which learns a kernel-wise hybrid binary and real-value soft mask for each task, while using the same backbone model. Such a soft mask can be viewed as a superposition of a binary mask and a properly scaled real-value tensor, which offers a richer representation capability without low-level kernel support to meet the objective of low hardware overhead. 
We validate KSM on multiple benchmark datasets against recent state-of-the-art methods (e.g. Piggyback, Packnet, CPG, etc.), which shows good improvement in both accuracy and training cost.
\end{abstract}

\begin{figure*}[t]
  \centering
  \includegraphics[width=\linewidth]{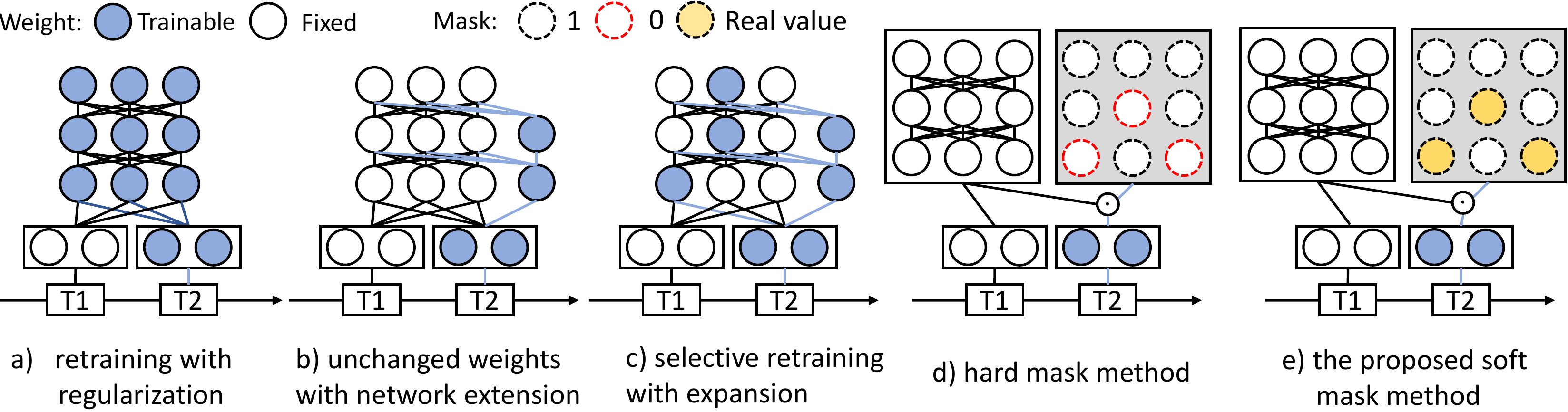}\\
\caption{
Overview of neural network approaches to overcome catastrophic forgetting, we consider the setting where each task retrains a new classifier. Except that, for the backbone model: a) retraining while
regularizing to prevent catastrophic forgetting with previously learned tasks; b) unchanged weights with network extension for representing new tasks;   c) selective retraining with possible
expansion; d) the hard mask method;  e) the proposed soft mask method.
}

\label{fig:overview}
\end{figure*}

\section{Introduction}
It is well-known that human and animals can learn new tasks without forgetting old ones. Nevertheless, conventional retraining of an existing Deep Neural Network (DNN) model for new tasks could easily result in the forgetting of old knowledge upon earlier tasks and thus degrades the performance. Such phenomenon is known as \textit{catastrophic forgetting}, which widely exists in continual learning~\cite{kirkpatrick2017overcoming}. We note that the continual learning refers that a model is incrementally updated over a sequence of tasks, performing knowledge transfer from earlier tasks to current one.

Recent works~\cite{li2017learning,kirkpatrick2017overcoming,riemer2018learning,chaudhry2018efficient,yoon2017lifelong,mallya2018packnet, hung2019compacting, yoon2019scalable, parisi2019continual} have made significant efforts in introducing various countermeasures to overcome the catastrophic forgetting issue.
As illustrated in~\cref{fig:overview}, given a well-trained model on the initial task, those countermeasures could be generally summarized into four methods: 
\textbf{a)} Training the model w.r.t the new task with regularization to prevent drastic weight update, thus maintaining the performance of old tasks~\cite{li2017learning,kirkpatrick2017overcoming,riemer2018learning,chaudhry2018efficient};
\textbf{b)} Freezing the backbone model of old tasks, while introducing additional task-specific weights for the new tasks~\cite{rusu2016progressive};
\textbf{c)} Selectively retraining partial weights of backbone model, as well as adding additional task-specific parameters on new tasks~\cite{mallya2018packnet, hung2019compacting, yoon2017lifelong, yoon2019scalable};
\textbf{d)} Fixing the backbone model weights and only learning a binary mask to select relevant weights for new tasks~\cite{mallya2018piggyback}. 
Overall, it can be seen that the trend is to introduce task-specific parameters while squeezing out model redundancy.

We elaborate further on the above four methods against catastrophic forgetting  (\cref{fig:overview}).  Method-a) could not effectively prevent the catastrophic forgetting with the growing number of new tasks.
In contrast, in Method-b) and Method-d), since the backbone model weights correspond to old tasks are frozen, the inference accuracy on old tasks is guaranteed.   However, Method-b) fails to achieve good performance on new tasks, owing to parameters correspond to old tasks (i.e., backbone model) are not effectively utilized.
More recently, the method-c), named compress-grow-based approach (CPG \cite{hung2019compacting}), handles the problem by iteratively compressing (via pruning) and then growing additional task-specific parameters. Note that, it will expand the model capacity until the accuracy on new task is maximized. Unfortunately, such method is at the cost of one order larger training time and computing resources, since it combines model pruning, weights selection, model channel expansion and even weight regularization. Although it largely alleviates the catastrophic forgetting and performs well on old and new tasks, such extremely high training cost, in terms of both training time and computing resources, makes it impossible to deploy into state-of-the-art popular edge based or mobile computing based continual learning domain. 

In this work, we propose a new training method called \textit{Kernel-wise Soft Mask} (KSM), which learns a kernel-wise hybrid binary and real-value soft mask for each task, while keeping the backbone model fixed. The KSM method has capability to mitigate catastrophic forgetting, to better transfer knowledge from old tasks, and more importantly, to maximize the training efficiency. 
More specifically, we use a similar network architecture  as Piggyback~\cite{mallya2018piggyback} (\cref{fig:overview}(d)), which introduces a mask tensor to perform weight re-factorization.
We want to highlight that our method differs from piggyback or other prior works in the following aspects:
\begin{enumerate}
    \item \textbf{Kernel-wise mask sharing.} To reduce the mask size and improve the computation efficiency in hardware, we design the mask in kernel-wise, instead of element-wise. For example, for a 3 by 3 kernel, the mask size would reduce by 9 times.
    \item \textbf{Soft mask.} 
    To boost the knowledge representation ability without involving additional training cost, we decompose the mask into a binary mask and partial real-value scaling coefficient tensor.
    \item \textbf{Softmax trick.} To eliminate gradient estimation in binary mask training, we propose to leverage softmax trick to relax the gradient calculation for mask during training. 
\end{enumerate}
Benefiting from these techniques, the proposed KSM method could achieve CPG-like (or even better) accuracy performance, while keeping Piggyback-like (or even better) training speed. 
It is also worth noting that, in this work, we only focus on KSM without growing the backbone model. But it is compatible with model expansion if needed, which will be investigated in future works.

    
    

\section{Related Work}

\subsection{Continual learning} 
The related works in continual learning can be categorized into network regularization and dynamic architecture.
\subsubsection{Network regularization}
Network regularization approaches aim to constrain updates of model weights by applying penalties to keep the learned tasks information. \cite{li2017learning} proposes Learning Without Forgetting (LWF), which shrink the prediction distance between current task and previous tasks by knowledge distillation~\cite{hinton2015distilling}.
EWC~\cite{kirkpatrick2017overcoming} uses Fisher’s information to evaluate the importance of weights for old tasks, and 
slow down the update of the important weights. Based on similar ideas, \cite{zenke2017temporal} alleviates catastrophic forgetting by allowing individual synapses to estimate their importance for solving a learned task.

\subsubsection{Dynamic architecture}
Network regularization methods can not completely guarantee to solve catastrophic forgetting, especially with unlimited number of tasks.
An another method to address this challenge is by dynamically expanding the network architecture. \cite{rusu2016progressive} proposes to expand the network by generating new sub-network with fixed size for each task, while fix the backbone model. \cite{yoon2017lifelong} selectively retrain the old network while expanding with limited neurons by group-sparsity regularization, and then splitting and duplicating the neurons to avoid catastrophic forgetting.
Beyond that, PackNet~\cite{mallya2018packnet} avoids this issue by identifying weights important for prior tasks through network pruning, while keeping the important weights fixed after training for a particular task.
In contrast to directly expanding model architecture, \cite{yoon2019scalable} adds additional task-specific parameters for each task and selectively learn the task-shared parameters together.
CPG~\cite{hung2019compacting} combines the model pruning, weight selection and model expansion methods, which gradually prunes the task-shared weights and then learns additional task-specific weights. Moreover, it could uniformly expand the model channels in each layer if the current accuracy can not meet the requirement. 
\begin{figure}
    \centering
    \includegraphics[width=0.9\linewidth]{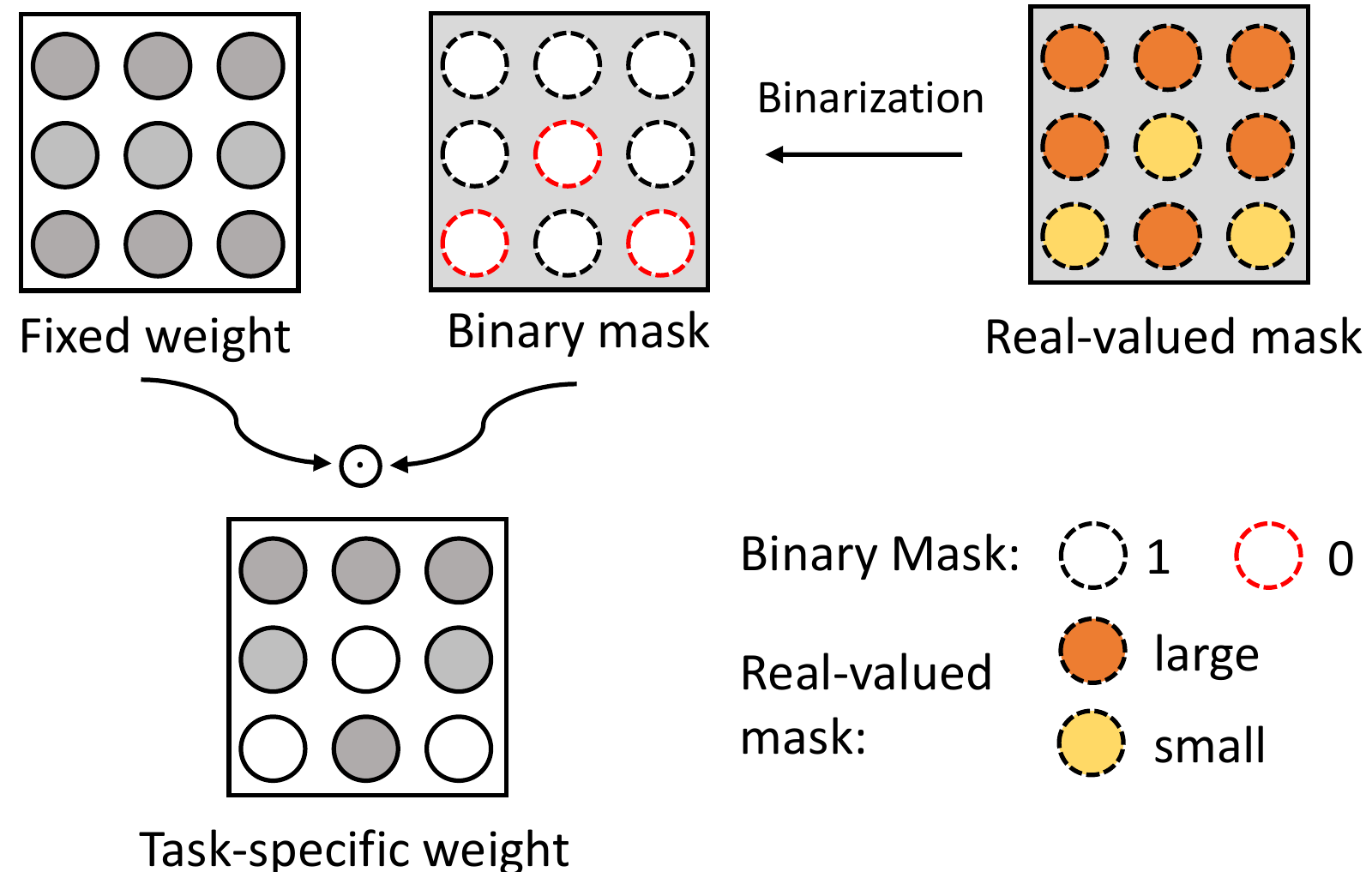}
    \caption{Illustration of Piggyback to learn a binary mask given a background model~\cite{mallya2018piggyback}. }
    \label{fig:piggyback}
\end{figure}
\subsection{Multi-domain learning}
Multi-domain learning~\cite{rebuffi2017learning, rosenfeld2018incremental} aims to build a model, which can adapt a task into multiple visual domains without forgetting previous knowledge, and meanwhile using as fewer parameters as possible. \cite{rosenfeld2018incremental} proposes to recombine the weights of the backbone model via controller modules in channel-wise. \cite{liu2019end} proposes  domain-specific attention modules for the backbone model. One of the most related method is Piggyback~\cite{mallya2018piggyback}, which solves the issue by learning task-specific binary masks for each task, as illustrated in~\cref{fig:overview}(d). They achieve this by generating the real-value masks which own the same size with weights, passing through a binarization function to obtain binary masks, that are then applied to existing weights. We denote the real-value mask and binary mask as $\tM^r$ and $\tM^b$ respectively, then, the binarization function is given by:
\begin{equation}
\label{eqt:binary}
    \textup{Forward}:~~~ \tM^{\textrm{b}} = 
    \begin{cases}
    1 & \textrm{if} ~\tM^{\textrm{r}} \geq \tau\\
    0 & \textrm{otherwise}
    \end{cases}
\end{equation}
\begin{equation}
    \textup{Backward}:~~~ \nabla \tM^{\textrm{b}} = \nabla \tM^{\textrm{r}}
\label{eqt:ste}
\end{equation}
Where $\tau$ is a constant threshold value. However, the gradient of binarization is non-differential during back-propagation. They use the Straight-Through Estimator (STE)~\cite{hubara2016binarized} to solve this problem, which estimates the gradient of real-value mask by the gradient of binary mask as shown in~\cref{eqt:ste}.

\begin{figure}[t]
  \centering
  \includegraphics[width=1.0\linewidth]{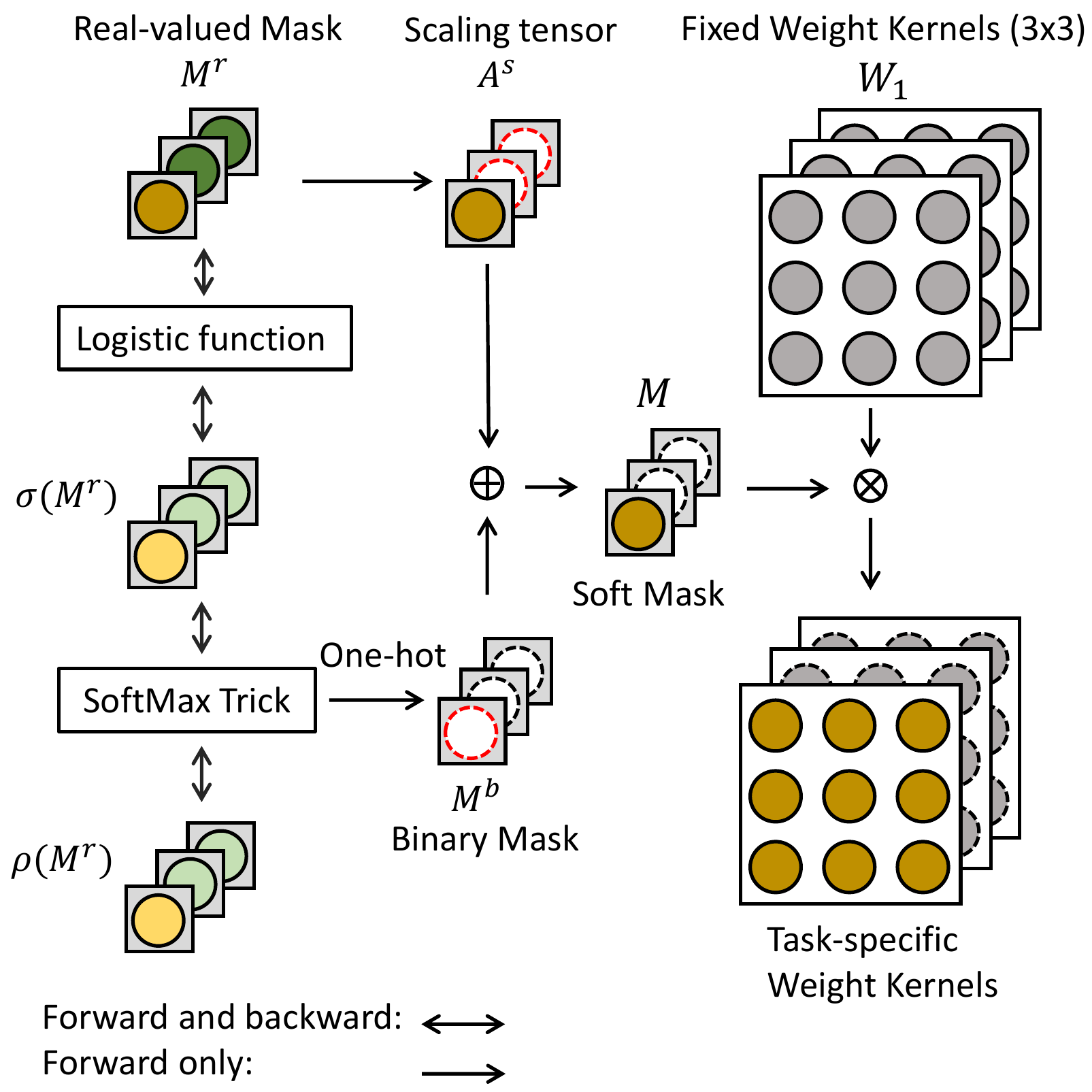}\\

\caption{
Overview of the proposed soft mask (KSM) learning method.
Give a task $t$, we aim to learn a task-specific soft mask $\tM_t$, by refactoring the fixed backbone weight to favor the current task. $\tM_t$ is decomposed into a binary mask $\tM^\textrm{b}_t$ and a scaling coefficient tensor $\tA^\textrm{s}_t$ (\cref{eqt:soft_mask}). To obtain $\tM^\textrm{b}_t$, the learnable real-value mask $\tM_t$ pass through a logistic function (\cref{eqt:sigmoid}) and a softmax function  (\cref{eqt:soft_trick}) successively. In addition, scaling tensor $\tM^\textrm{b}_t$ is generated by $\tM^\textrm{r}_t$ (\cref{eqt:scaling}). 
During training backward, the real-value mask can be updated without gradient estimation. After training, only the soft mask is saved for testing.   
}
\label{fig:soft_mask}
\end{figure}

\section{Kernel-wise Soft Mask Method}


Different from the conventional multi-task learning where the data of all tasks is available at training time, we consider a continual learning setting in which new tasks ($\{\mathcal{T}_1, \mathcal{T}_2, ..., \mathcal{T}_N\}$) arrive sequentially and cannot be used for training future tasks.
Given a convolution layer, we denote the weights $\tW^{(l)} \in \R^{c_\textrm{in}\times c_\textrm{out}\times kh\times kw}$, where $c_\textrm{in}, c_\textrm{out}, kh, kw$ refers the weight dimension of $l$-th layer, including \#output channel, \#input channel, kernel height and width, respectively. 
We denote the dataset of the $t$-th task ($\mathcal{T}_t$) as $\tD_t = \{\vx_t, \vy_t\}$, where $\vx_t$ and $\vy_t$ are vectorized input data and label pair.
To adapt the pre-trained backbone model with the parameter $\{\tW_1\}$ from the initial task $\mathcal{T}_1$ to a new task $\mathcal{T}_t$, we intend to learn a task-specific soft mask $\tM_t\in \mathbb{R}$ that is applied to the fixed parameter $\tW_1$ to provide good performance.
Based on this idea, the optimization objective can be mathematically formalized as:
\begin{equation}
\min_{\tM_t} {\mathcal{L}}{ \Big ( }f ( \vx_t; \{\tM_t \times \tW_{1}\}), \vy_t {\Big )}
\label{eqt:loss}
\end{equation}
As described in~\cref{eqt:loss}, the inference performs matrix multiplication between mask tensor ${\tM}$ and weight, thus refactorizing the weights to favor the new tasks. Such mask based method differs from prior mask-based counterparts \cite{mallya2018piggyback} in following aspects:
\begin{enumerate}
\item \textbf{Kernel-wise Mask Sharing}. Since the task-specific weights are refactorized from the backbone model via the task-specific mask, the size of mask directly determines the computation and model overhead for domain adaption objective. 
Instead of utilizing the mask in the identical size with weights as in~\cite{mallya2018piggyback}, we introduce the compact mask where each mask element is shared by the kernel $kh \times kw$. Such kernel-wise mask method properly alleviates the computation and memory burden, with potential to outperform the existing methods in terms of accuracy.


\item \textbf{Soft Mask.} In contrast to prior works leveraging binary mask ($\tM_t \in \{0,1\}$), we use the real-value mask ($\tM_t \in \mathbb{R}$) instead (aka. soft mask) with sparse patterns. 
Note that  our method still includes sparse elements as the binary counterpart, but the zero elements are replaced with real values.
Such a soft mask can be viewed as a superposition of a binary mask $\tM^b_t$ and  a scaling coefficient tensor $\tA^s_t$, which can be expressed as:
\begin{equation}
\tM_t = \tM^b_t  +  \tA^s_t.
\label{eqt:soft_mask}
\end{equation}
The above modification empowers the soft mask with a richer representation capability, without low-level kernel support to meet the objective of low hardware overhead.

\item \textbf{Softmax trick for better gradient calculation.}
Since the soft mask above includes sparse portion, there still exists the non-differential issue. Instead of utilizing Straight-Through Estimator (STE) in the binary mask counterpart, we propose to leverage the softmax trick to relax the categorical objective. Compared to the STE method, the softmax trick could provide better gradient calculation, to achieve higher accuracy on new tasks.

    
\end{enumerate}

  \cref{fig:overview}depicts  the evolution from prior implementation to our method. More details of our soft mask based method are presented in the following subsections.


\subsection{Soft mask}

As in Piggyback method~\cite{mallya2018piggyback}, the adopted binary mask compulsively divides the background model into two categories: task-relevant or -irrelevant, represented by 1 or 0 in the binary mask respectively. 
To better leverage the knowledge of the background model, we introduce an additional trainable real-value scale coefficient tensor $\tA^\textrm{s}$ as a replacement of the zero elements in the binary mask counterpart. In this way, it can  improve the learning capacity without time-consuming re-training of zeroed-out weights in CPG~\cite{hung2019compacting}.
Next, we seek to answer the following two key questions:
\begin{itemize}
\item How to generate the scaling coefficient tensor without involving additional training parameters or cost?

\item Where to apply these scaling factors?
\end{itemize}
Intuitively, the trainable soft mask should be utilized to represent the relevant or importance levels w.r.t to the corresponding weight kernel of the background model. In light of this, we propose to directly use it as the scaling factor. 
In practice, the magnitude of values in the real-value mask is typically very small (i.e. 0.01), even with negative values. 
In our method, we normalize those values to treat them as the scaling factor of each kernel when learning new tasks.
Next, we  apply those normalized scaling factors only to the kernels that are zeroed-out in the binary mask, so as to create a soft mask and avoid mask size increasing significantly due to those real values.  
As shown in Fig.~\ref{fig:soft_mask}, the above two steps can be achieved by inverting `0' and `1' in the generated binary mask $\tM^b$, followed by  multiplying with the real-value mask $\tM^r$. 
The scaling factor is given by:
\begin{equation}
      \tA^s = \overset{\sim}\tM{^b} \cdot \textrm{normal}(\tM_t^r.\textrm{detach()}) 
\label{eqt:scaling}
\end{equation}
Where $\overset{\sim}\tM{^b}$ inverts 0 and 1 in the $\tM{^b}$. The `detach' is used to only grasp the values of $\tM{^r}$ without influence the backpropogation. Note that, since all the masks are set in kernel-wise, each mask value will be applied on a kernel weight as shown in~\cref{fig:soft_mask}.

In short, we generate the soft mask $\tM$ by combining the binary mask $\tM^b$ and the scaling factor $\tA^s$ as shown in~\cref{eqt:soft_mask}. It can be understood as we fix the important kernels (`1' in binary mask) and scale the unimportant kernels (`0' in binary mask) to be different trainable levels for the new task. The soft mask is generated in this way, mainly for the following two reasons:
\begin{enumerate}
    \item Directly utilizing the already existing real-value mask does not involve additional trainable parameters or changing the backbone model architecture, indicating that it can be trained with no extra cost.
    
    \item These scaling factors increase the model capacity for the new task, with very small mask size increase due to the facts that 1) real-values occupy a small portion in the mask and 2) our kernel-wise mask dimension is already much smaller than traditional element-wise mask. We will quantify the overhead and the sparsity level in the analysis later. 
\end{enumerate}

\subsection{Softmax trick for gradient calculation} 

\cite{mallya2018piggyback} proposes a masking method, where they train a real-value mask followed by a hard threshold function to binarize the mask as depicted in~\cref{eqt:binary}.
However, the binarization function is not differentiable, the general solution is to skip the threshold function during backpropagation and update the real mask by directly using gradients computed from binary mask, which is known as Straight Through Estimator (STE) as shown in Eq.\ref{eqt:ste}. Different from that, we propose a method to eliminate the gradient estimation step and make whole mask learning compatible with existing gradient based backpropagation training process.

First, we relax the binarization function in Eq.\ref{eqt:binary} to a continuous logistic function:
\begin{equation}
    \sigma(\tM^r) = \frac{1}{1+\textrm{exp}(-k(\tM^r-\tau))}
\label{eqt:sigmoid}
\end{equation}
where $k$ is a constant. 
Note that the logistic function becomes closer to hard thresholding function when k is increasing. 
The partial derivative of the logistic function is:
\begin{equation}
    \frac{\partial \sigma(\tM^r)} {\partial \tM^r} = k \cdot \sigma(\tM^r) \cdot (1-\sigma(\tM^r))
\end{equation}
In this work, we treat $\sigma(\tM^r)$ as a probability mask to estimate the importance level of the corresponding weight kernels to save training time without involving extra parameters.

When considering it as a probability mask, sampling from a Bernoulli distribution is a reasonable and popular way to generate, but such sampling procedure is not differentiable. To overcome this issue, we leverage the softmax trick, which performs a differential sampling to approximate a categorical random variable. Summarizing, we define $p(\cdot)$ using the softmax trick as
\begin{equation}
    p(\tM^r) = \frac{ \textrm{exp}((\textrm{log} \pi_0)/T)}{\textrm{exp}((\textrm{log}\pi_0)/T) + \textrm{exp}((\textrm{log}\pi_1)/T)}
\label{eqt:soft_trick}
\end{equation}
Where $\pi_0$ and $\pi_1$ represent $1-\sigma(\tM^r)$ and $\sigma(\tM^r)$ respectively. The temperature $T$ is a hyper-parameter to adjust the range of input values, meanwhile choosing larger one could avoid gradient vanishing during back-propagation. Note that the output of $p(\tM^r)$ closer to a Bernoulli sample as $T$ towards to 0. 

Benefiting from the differentiable property of~\cref{eqt:sigmoid} and \cref{eqt:soft_trick}, the real-value mask $\tM^r$ can be embedded with existing gradient based backpropagation training without gradient approximation. During training, most values in the distribution of $p(\tM^r)$ will move towards either 0 and 1. To represent $p(\tM^r)$ as binary format, we use the one-hot code of $p(\tM^r)$ during training forward, which has no influence on the real-value mask to be updated for back-propagation. 



In the end, the soft mask is generated as described in~\cref{eqt:soft_mask}. During forward, the input-output relationship of one layer is given by
$\vy = \tW_{1} \cdot (\tM^\textrm{b}  +  \tA^\textrm{s})\vx$. According to the chain rule in the back-propagation, the gradient of such soft mask is given by:
\begin{equation}
\begin{gathered}
    \nabla \tM^s = (\frac{\partial E} {\partial \vy}) \cdot (\frac{\partial \vy} {\partial p(\tM^r)}) \cdot (\frac{\partial p(\tM^r)} {\partial \sigma(\tM^r)}) \cdot (\frac{\partial \sigma(\tM^r)} {\partial \tM^r}) \\
\end{gathered}
\end{equation}
Where the partial derivative of each term is given by: 
\begin{equation}
\begin{gathered}
    \frac{\partial E} {\partial \vy} = \nabla \vy \\
    \frac{\partial \vy} {\partial p(\tM^r)} = \vx^T \cdot \tW_1 \\
    \frac{\partial p(\tM^r)} {\partial \sigma(\tM^r)} = - \frac{p(\tM^r)(1-p(\tM^r))} {T\sigma(\tM^r)(1-\sigma(\tM^r))}
\end{gathered}
\end{equation}

By doing so, the proposed method can optimize the soft mask in an end-to-end manner, where every step is differentiable. We illustrate the complete algorithm in Algorithm \ref{algo:1}. During training, we save the optimized $\tM^*$, and then directly applying it to the corresponding weight for testing.


\setlength{\textfloatsep}{0pt}
\begin{algorithm}
\caption{The proposed soft mask learning}\label{algo:1}
\begin{algorithmic}[1]
\Require{Give the initial task $\mathcal{\tau}_1$ and the backbone model with the parameter $\tW_1$, the threshold $\tau$ and temperature $T$ }
\For{Task t = 2, ..., N}
    \State Get data $\vx_t$ and label $\vy_t$
    \State $\tM_t^b \leftarrow$ one-hot$(p(\tM_t^r))$
    \State $\overset{\sim}\tM{^b_t}\leftarrow \textrm{invert}(\tM_t^b)$
    \State $\tM_t \leftarrow \tM^b_t  + \overset{\sim}\tM{^b_t} \cdot \textrm{normal}(\tM_t^r.\textrm{detach}()) $
    \State $\tM^*_t \leftarrow \min_{\tM^s_t} {\mathcal{L}}{ \Big ( }f ( \vx_t;\tW_{1} \cdot \tM_t), \vy_t {\Big )}$
    \State During testing, execute $f ( \vx_t;\tW_{1} \cdot \tM^*_t)$
\EndFor
\end{algorithmic}
\end{algorithm}

\begin{table*}[t]
\caption{The accuracy (\%) and training cost (s) on Twenty Tasks of CIFAR-100. Considering those accuracy and training time comparison, we could achieve best average accuracy and around $\sim10\times$ faster than CPG.}
\scalebox{0.65}{
\begin{tabular}{cccccccccccccccccccccc|c}
\hline
Methods &  & 1 & 2 & 3 & 4 & 5 & 6 & 7 & 8 & 9 & 10 & 11 & 12 & 13 & 14 & 15 & 16 & 17 & 18 & 19 & 20 & \textbf{Avg} \\ \hline
\multirow{2}{*}{PackNet} & Acc & 66.4 & 80.0 & 76.2 & 78.4 & 80.0 & 79.8 & 67.8 & 61.4 & 68.8 & 77.2 & 79.0 & 59.4 & 66.4 & 57.2 & 36.0 & 54.2 & 51.6 & 58.8 & 67.8 & 83.2 & 67.5 \\ \cline{2-23} 
 & \multicolumn{1}{l}{Time} & \multicolumn{1}{l}{334} & \multicolumn{1}{l}{360} & \multicolumn{1}{l}{370} & \multicolumn{1}{l}{379} & \multicolumn{1}{l}{382} & \multicolumn{1}{l}{385} & \multicolumn{1}{l}{385} & \multicolumn{1}{l}{389} & \multicolumn{1}{l}{234} & \multicolumn{1}{l}{358} & \multicolumn{1}{l}{370} & \multicolumn{1}{l}{378} & \multicolumn{1}{l}{384} & \multicolumn{1}{l}{385} & \multicolumn{1}{l}{384} & \multicolumn{1}{l}{337} & \multicolumn{1}{l}{359} & \multicolumn{1}{l}{371} & \multicolumn{1}{l}{377} & \multicolumn{1}{l|}{382} & \multicolumn{1}{l}{365} \\ \hline
\multirow{2}{*}{Piggyback} & Acc & 65.8 & 78.2 & 76.4 & 79.8 & 86.0 & 81.0 & 79.4 & 82.4 & 81.8 & 86.4 & 87.8 & 76.0 & 82.8 & 80.6 & 48.2 & 70.4 & 65.0 & 71.80 & 87.80 & 90.6 & 77.1 \\ \cline{2-23} 
 & \multicolumn{1}{l}{Time} & \multicolumn{1}{l}{100} & \multicolumn{1}{l}{150} & \multicolumn{1}{l}{102} & \multicolumn{1}{l}{113} & \multicolumn{1}{l}{154} & \multicolumn{1}{l}{102} & \multicolumn{1}{l}{121} & \multicolumn{1}{l}{119} & \multicolumn{1}{l}{97} & \multicolumn{1}{l}{130} & \multicolumn{1}{l}{84} & \multicolumn{1}{l}{110} & \multicolumn{1}{l}{96} & \multicolumn{1}{l}{120} & \multicolumn{1}{l}{106} & \multicolumn{1}{l}{97} & \multicolumn{1}{l}{97} & \multicolumn{1}{l}{106} & \multicolumn{1}{l}{110} & \multicolumn{1}{l|}{119} & \multicolumn{1}{l}{111} \\ \hline
\multirow{2}{*}{CPG} & Acc & 66.6 & 76.2 & 78.2 & 80.6 & 86.4 & 83.0 & 81.4 & 82.4 & 82.0 & 86.8 & 86.8 & 81.4 & 82.8 & 82.0 & 50.4 & 72.4 & 66.2 & 71.2 & 85.6 & 91.6 & 78.7 \\ \cline{2-23} 
 & Time & 629 & 2101 & 2123 & 2120 & 2121 & 2127 & 2116 & 2120 & 2122 & 2121 & 2122 & 2115 & 2127 & 2125 & 2126 & 2114 & 2124 & 2126 & 2123 & 2125 &  2046 \\ \hline
\multirow{2}{*}{Ours} & Acc & 67.2 & 78.0 & 78.8 & 78.4 & 85.6 & 82.6 & 80.2 & 83.4 & 82.6 & 89.4 & 88.4 & 80.6 & 83.2 & 80.8 & 52.8 & 73.2 & 67.8 & 72.6 & 88.0 & 92.0 & \textbf{79.2} \\ \cline{2-23} 
 & Time & 130 & 81 & 111 & 123 & 123 & 127 & 62 & 106 & 88 & 78 & 95 & 85 & 73 & 88 & 90 & 90 & 80 & 95 & 96 & 65 & \textbf{94.3} \\ \hline
\end{tabular}}
\label{tab:cifar100_20}
\end{table*}

\section{Experiments}
\subsection{Datasets and backbone architectures}

Similar as prior works, we use VGG16-BN~\cite{simonyan2014very} and ResNet50~\cite{he2016deep} as the backbone architectures for the following datasets:

\noindent\textbf{ImageNet-to-Sketch}
In this experiments, six image classification datasets are used: CUBS~\cite{wah2011caltech}, Stanford Cars~\cite{krause20133d}, Flowers~\cite{nilsback2008automated}, WikiArt~\cite{saleh2015large} and Sketch~\cite{eitz2012humans}. We use the ResNet50 as the backbone model which are trained on ImageNet dataset~\cite{russakovsky2015imagenet}, then fine-tunes the fine-grained datasets sequentially.
    
\noindent\textbf{Twenty Tasks of CIFAR-100} We divide the CIFAR-100 dataset into 20 tasks. Each task has 5 classes, 2500 training images, and 500 testing images. In the experiment, VGG16-BN model (VGG16 with batch normalization layers) is employed to train the 20 tasks sequentially.

\subsection{Competing Method to be Compared}
To test the efficacy of our method, we compare it with recent several representative methods in three categories:
\begin{itemize}
    \item \textbf{Whole model fine-tuning:} Fine-tuning the whole model for each task individually
    \item \textbf{PiggyBack~\cite{mallya2018piggyback}} It fixes the backbone weights and then learns a binary mask to select partial weights for new tasks.
    \item \textbf{PackNet~\cite{mallya2018packnet}} It first prunes unimportant weights, and then fine-tunes them for learning new tasks.
    \item \textbf{CPG~\cite{hung2019compacting}} It combines PackNet and PiggyBack to gradually prune, pick and grow the backbone model for learning new tasks sequentially.
\end{itemize}

\subsection{Results on ImageNet-to-Sketch dataset}
In this experiment, following the same settings in the works of CPG ~\cite{hung2019compacting} and Piggyback~\cite{mallya2018piggyback}, We train each task for 30 epochs using the Adam optimizer. The initial learning rate is 1e-4, which is decayed by a factor of 10 after 15 epochs. 

\subsubsection{Accuracy comparison}
The accuracy of the five classification tasks is tabulated in~\cref{tab:imagenet}. 
For the first ImageNet task, CPG and PackNet perform slightly worse than the others, since both methods have to compress the model via pruning. 
Then, for the following five fine-grained tasks, the proposed method could achieve all better accuracy comparing with Piggyback and PackNet. 
Even comparing with the individually fine-tuning whole model for each task, we could still achieve better performance except WikiArt dataset.
In comparison to CPG that requires one order more training time (\cref{fig:train_cost_imagenet}), our method achieves better accuracy in tasks of CUBS, Flowers and Sketch. However, we admit that, owing to a small portion of real-values in the mask, our method needs slightly more model size than other methods. 
Note that, the model size reported in~\cref{tab:imagenet} includes both the backbone model and introduced mask size.

\subsubsection{Training time comparison} 
\begin{figure}
    \centering
    \includegraphics[width=\linewidth]{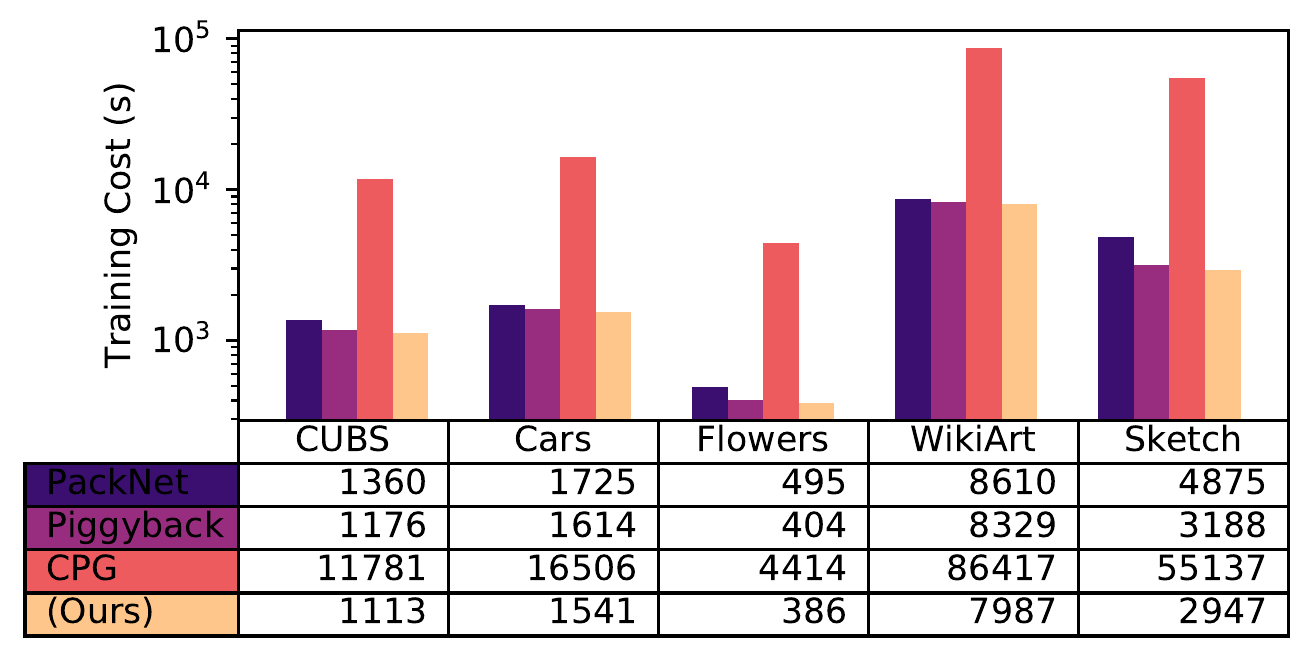}
    \caption{Training cost on ImageNet-to-Sketch datasets with various continual learning methods.}
    \label{fig:train_cost_imagenet}
\end{figure}

To do a fair comparison, all the methods are trained on the single Quadro RTX 5000 GPU with the same batch size. 
\cref{fig:train_cost_imagenet} summarizes the whole training time for each method. 
First, our method slightly outperforms Piggyback, since the proposed soft mask learning method (as illustrated in~\cref{eqt:sigmoid} and \cref{eqt:soft_mask}) is faster than the binarization function in real hardware implementation. Then, ours and Piggyback could both achieve better speed than PackNet, since PackNet needs to retrain weights which is slower than training a mask. Last, it is very obvious that CPG requires more than $\sim10\times$ more training time than all rest methods, while 3 out of 5 tasks have lower accuracy than ours as shown in \cref{tab:imagenet}.


\begin{table}[h]
\caption{Accuracy on ImageNet-to-Sketch dataset}
\scalebox{0.85}{
\begin{tabular}{cccccc}
\hline
Dataset & \multicolumn{1}{l}{Finetune} & \multicolumn{1}{l}{PackNet} & \multicolumn{1}{l}{Piggyback} & \multicolumn{1}{l}{CPG} & \multicolumn{1}{l}{Ours} \\ \hline
ImageNet &  & 75.71 & 76.16 & 75.81 & 76.16 \\
CUBS & 82.83 & 80.41 & 81.59 & 83.59 & \textbf{83.81} \\
Cars & 91.83 & 86.11 & 89.62 & \textbf{92.80} & 92.14 \\
Flowers & 96.56 & 93.04 & 94.77 & 96.6 & \textbf{96.94} \\
WikiArt & 75.60 & 69.40 & 71.33 & \textbf{77.15} & 75.25 \\
Sketch & 80.78 & 76.17 & 79.91 & 80.33 & \textbf{81.12}\\ \hline\hline
\begin{tabular}[c]{@{}c@{}}Model Size\\     (MB)\end{tabular} & 554 & 115 & 121 & 121 & 146  \\ \hline
\end{tabular}}
\label{tab:imagenet}
\end{table}

\begin{table*}[t]
\centering
\caption{The accuracy on Twenty Tasks of CIFAR-100 with different initial tasks. The accuracy of individual task on these five settings is slightly different.
Neverthless, the average accuracy is better than PackNet and Piggyback. Comparing with CPG, better accuracy could be achieved on three different initial types. }
\scalebox{0.75}{
\begin{tabular}{ccccccccccccccccccccc|c}
\hline
Initial & 1 & 2 & 3 & 4 & 5 & 6 & 7 & 8 & 9 & 10 & 11 & 12 & 13 & 14 & 15 & 16 & 17 & 18 & 19 & 20 & \textbf{Avg} \\ \hline
1 & 67.2 & 78.0 & 78.8 & 78.4 & 85.6 & 82.6 & 80.2 & 83.4 & 82.6 & 89.4 & 88.4 & 80.6 & 83.2 & 80.8 & 52.8 & 73.2 & 67.8 & 72.6 & 88.0 & 92.0 & 79.2 \\ \hline
5 &  67.0 & 77.2 & 77.6 & 79.2 & 84.8 & 82.6 & 78.0 & 85.2 & 82.8 & 88.8 & 88.4 & 80.8 & 84.2 & 81.4 & 50.2 & 71.8 & 67.0 & 71.2 & 86.0 & 91.8 & 78.8 \\ \hline
10 & 67.8 & 77.2 & 76.6 & 79.4 & 82.8 & 81.6 & 80.8 & 83.4 & 82.0 & 88.6 & 88.2 & 81.2 & 85.0 & 80.2 & 53.4 & 73.8 & 68.6 & 74.4 & 87.2 & 91.2 & \textbf{79.3} \\ \hline
15 &  67.6 & 78.2 & 77.0 & 77.0 & 81.8 & 82.6 & 78.4 & 83.4 & 83.2 & 86.6 & 88.4 & 80.0 & 83.0 & 78.0 & 51.2 & 70.8 & 67.8 & 67.8 & 86.4 & 91.0 & 78.0 \\ \hline
20 &  66.8 & 75.6 & 77.2 & 76.6 & 85.4 & 81.0 & 79.0 & 84.0 & 82.2 & 87.4 & 86.4 & 79.0 & 83.8 & 80.4 & 49.0 & 70.8 & 66.4 & 72.0 & 88.2 & 93.6 & 78.2\\ \hline
\end{tabular}}
\end{table*}

\subsection{Results on twenty tasks of CIFAR-100}
Different from the ImageNet-to-Sketch setting that relies on a pre-trained model on ImageNet dataset, in this experiment, we first train a task from scratch as the backbone model. Afterward, we fix the backbone model weights and learn the task-specific mask for the rest tasks sequentially. To conduct a fair comparison, we follow the same configuration as CPG, and select the same task as the initial task-1. 
Note that, since this work only focuses on continual learning without model expansion, all the CPG results are without expansion based on their open source code.

\subsubsection{Accuracy and training time comparison} Similar phenomenon can be observed with the ImageNet-to-Sketch setting.
Table.\ref{tab:cifar100_20} shows the accuracy and training cost for these methods. Our method could achieve completely better results than Piggyback and PackNet. In addition, comparing with CPG, we also could achieve better results in most tasks. In terms of training time, our method is around $\sim10\times$ faster than CPG. 

Considering those accuracy and training time comparison, it shows our method could achieve a well-performed knowledge transfer based on a weak backbone model which only trains on 2 classes. It is worthy to note that the initial task indeed influences the performance of rest tasks, since we fix the backbone weights all the time. In the next section, we will show that, even with different initial tasks, in all cases, our method could learn a mask that achieves good knowledge representation for new task.

\begin{figure}[h]
    \centering
    \includegraphics[width=\linewidth]{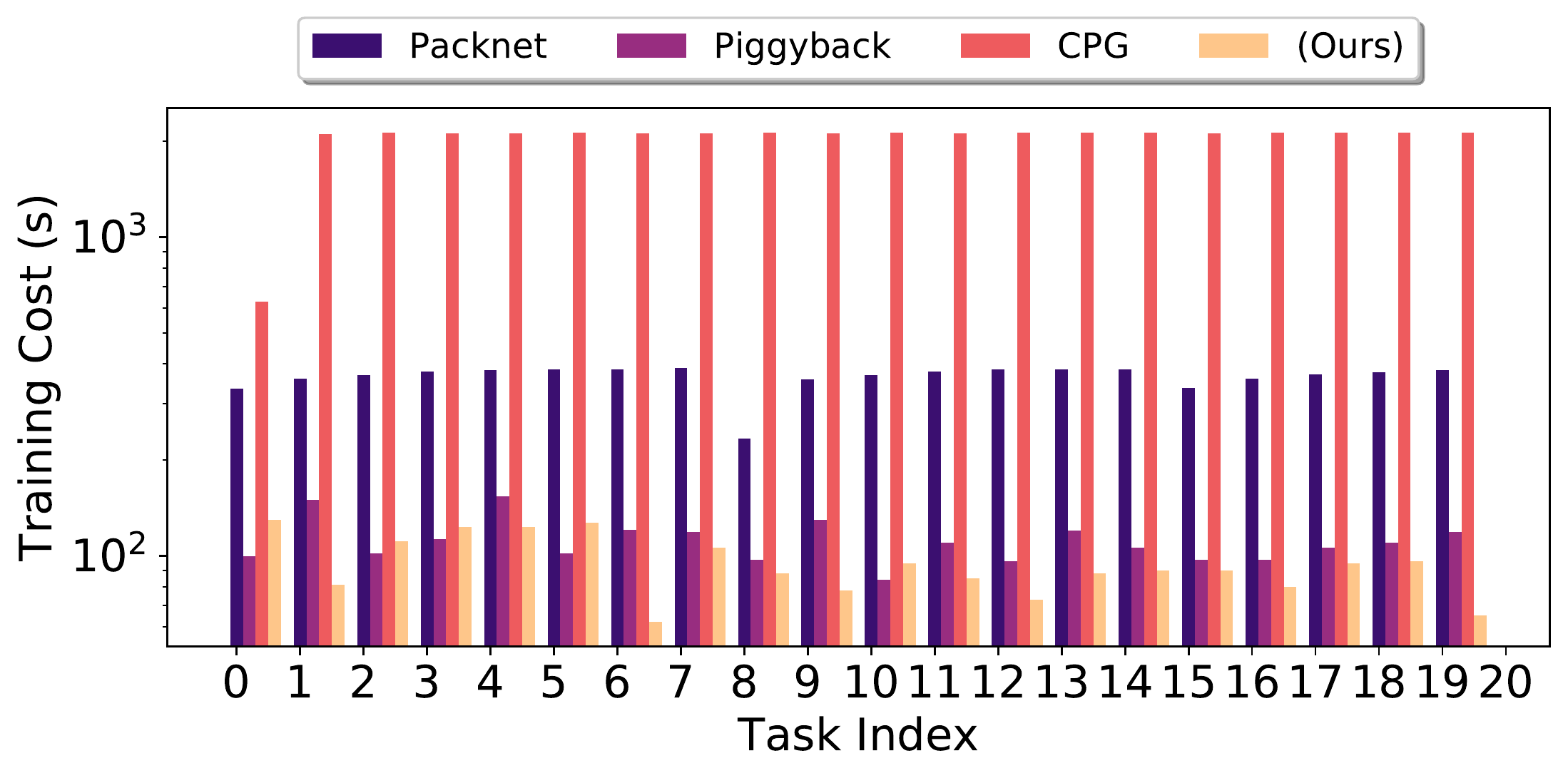}
    \caption{Training cost on  twenty tasks of CIFAR-100 with various continual learning methods.}
    \label{fig:init_tasks}
\end{figure}
\begin{figure}[t]
    \centering
    \includegraphics[width=0.95\linewidth]{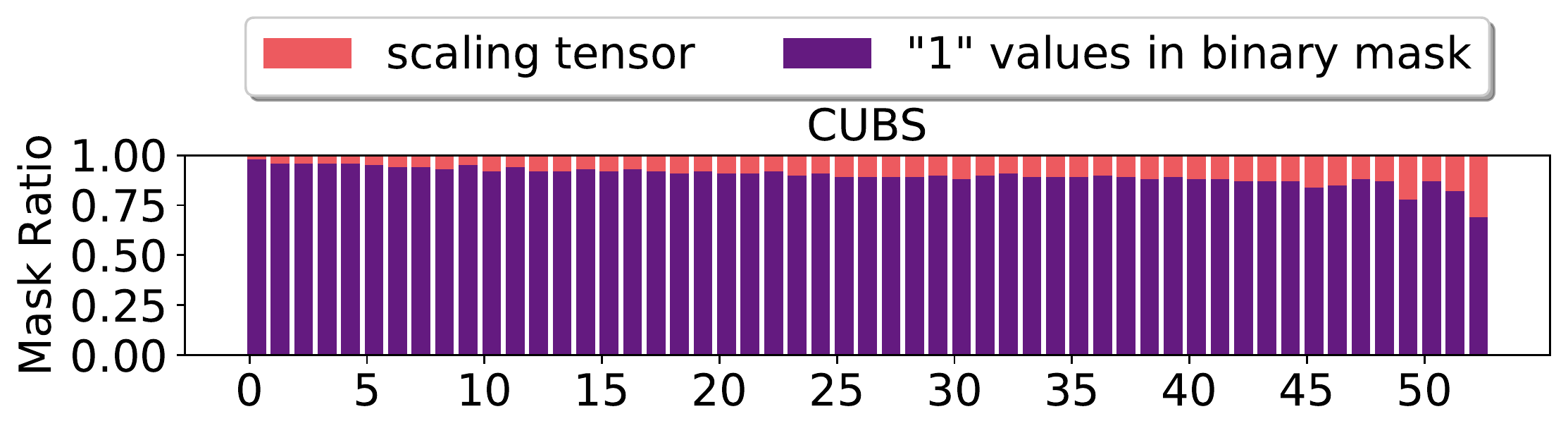}\\
    \includegraphics[width=0.95\linewidth]{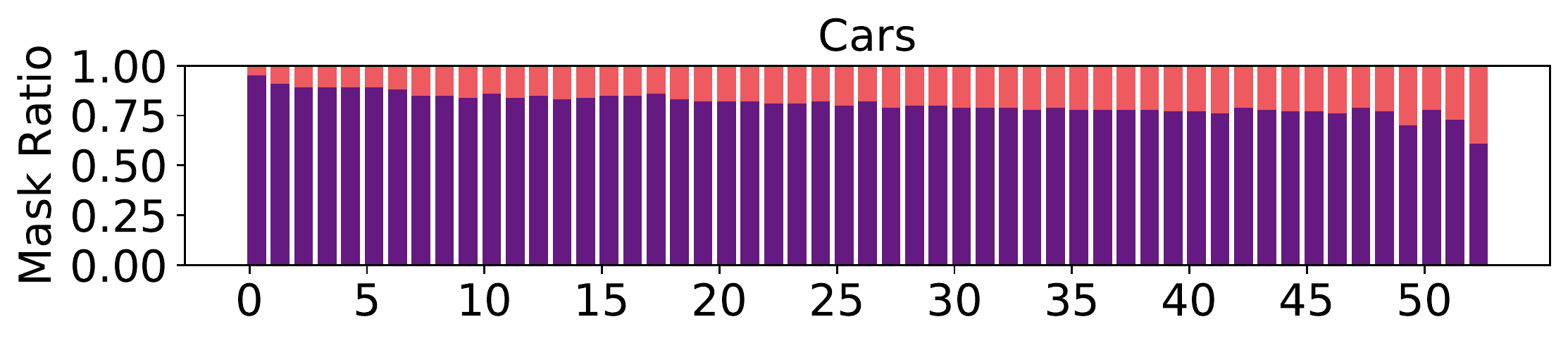}\\
    \includegraphics[width=0.95\linewidth]{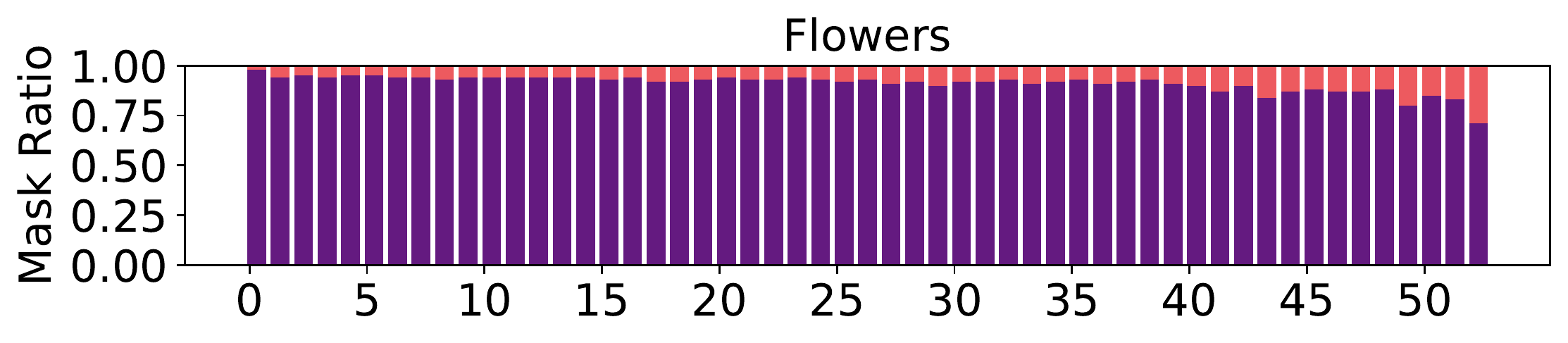} \\
    \includegraphics[width=0.95\linewidth]{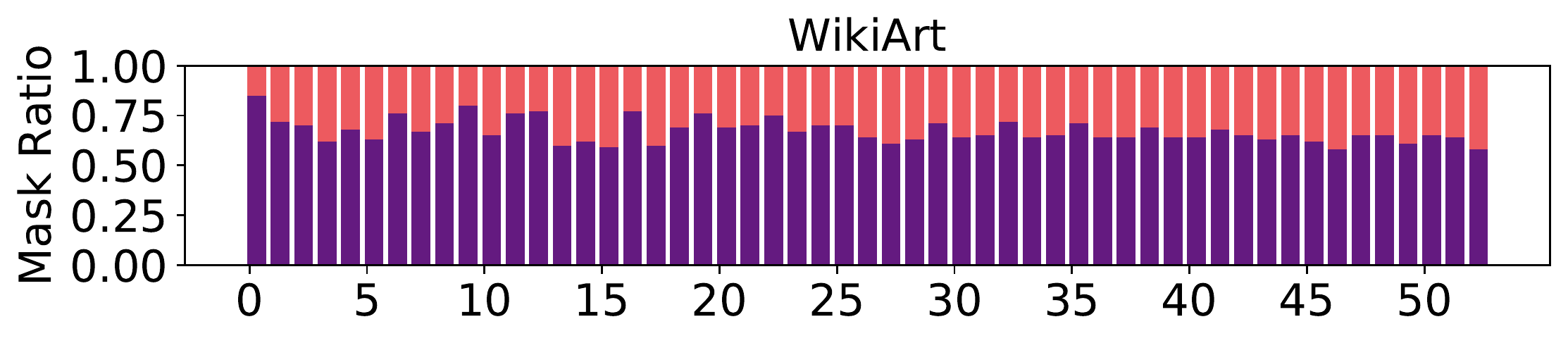}\\
    \includegraphics[width=0.95\linewidth]{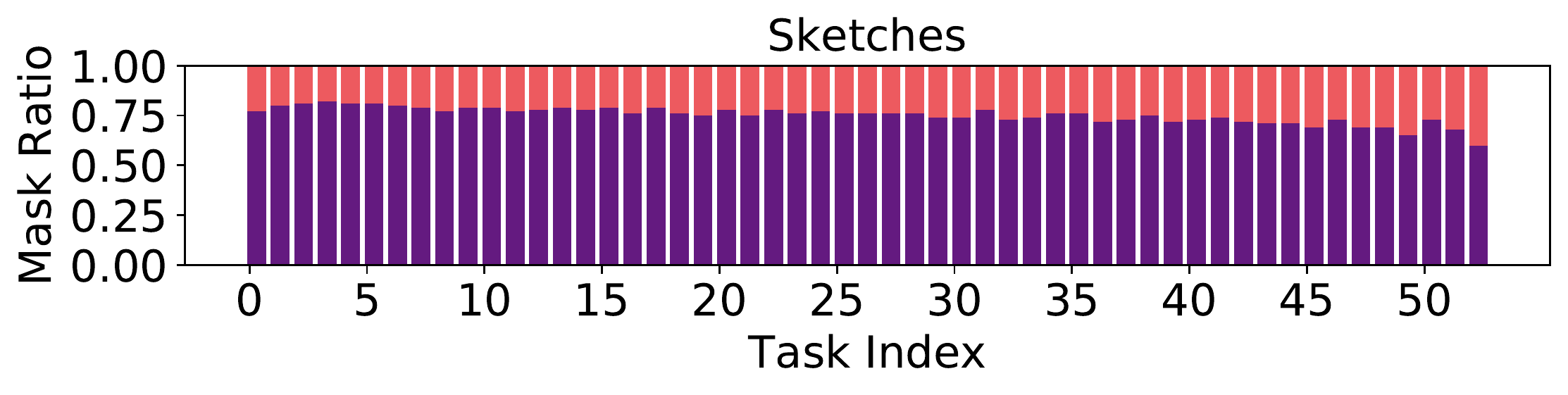}
    \caption{The ratio of two mask types visualization on ResNet50 for ImageNet-to-Sketches dataset.}
    \label{fig:arch_visual}
\end{figure}

\subsection{Ablation Study and Analysis}
\subsubsection{Kernel-wise, soft mask and softmax trick}
We study the individual effect of the three main techniques of our proposed method on ImageNet-to-Sketch dataset setting. As shown in~\cref{tab:Ablation}, we denote the `Piggyback-Soft' as replacing the 0 values in piggyback's binary mask with scaling factors, and denote the `Ours-softmax' as we only use the proposed softmax trick to generate binary mask. Also, we name the `Ker-wise' and `Ele-wise' as kernel-wise and element-wise mask respectively. The `Ours-Softmax' achieves better results than Piggyback, which means the proposed completely differentiable mask learning process with softmax trick could generate better optimization, since we don't have gradient estimation. In addition, `Piggyback-Soft' achieves better results than `Piggyback' proving that adding scaling factors to zeroed-out weights indeed improves the task-specific representation ability. Also, changing the mask to kernel-wise almost has very minor or neglectable influence for performance. In the end, the `Ours-Full' combines all three techniques, showing best overall performance in all datasets. 
\begin{table}[h]
\caption{The ablation study on the proposed method}
\label{tab:Ablation}
\scalebox{0.8}{
\begin{tabular}{cccccc}
\hline
Method & CUBS & Cars & Flowers & Wikiart & Sketch \\ \hline
Piggyback & 81.59 & 89.62 & 94.77 & 71.33 & 79.91 \\
Piggyback - Ker-wise & 81.76 & 89.57 & 94.88 & 70.30 & 79.95 \\
Piggyback - Soft & 82.26 & 91.17 & 95.85 & 73.12 & 80.22 \\
Ours - Softmax & 82.86 & 91.71 & 96.67 & 74.06 & 80.70  \\ 
Ours - Ele-wise & 83.79 & 92.18 & 96.90 & 75.0 & 81.10  \\ \hline
Ours - Full & 83.81 & 92.14 & 96.94 & 75.25 & 81.12 \\ \hline
\end{tabular}}
\end{table}

\subsubsection{The Effect of Different Initial Tasks}
\label{sec:init_task}
Different from the ImageNet-to-Sketch dataset setting that heavily relies on a strong pre-trained model, we randomly select a task and then train it from scratch as the initial model in Twenty Tasks of CIFAR-100 setting. 
To explore how does the initial task affects the performance of rest tasks, we randomly select five different tasks as the initial task as shown in~\cref{fig:init_tasks}. Thus, the accuracy of these five settings on each individual task is slightly different, since they own different domain shift levels. 
Nevertheless, the average accuracy is better than PackNet and Piggyback. Comparing with CPG, better accuracy could be achieved on three different initial types, which indicates that the proposed method could balance the domain shift with different initial tasks.

\subsubsection{Architecture and Soft Mask Visualization}
\cref{fig:arch_visual} visualizes the ratio of `1' values in binary mask and the scaling factor. Two observations can be found crossing all tasks: 1) Within a task, high-level layers need more changes than low-level layers, especially the last convolutional layer. 2) The scaling factor ratio seems can reflect the domain shift difficulty, for example, the largest dataset WikiArt need more changes than the smallest dataset Flowers.  

    
    

\section{Conclusion}
In this work, We propose a novel kernel wise soft mask method for multiple task adaption in the continual learning setting, which learns a hybrid binary and real-value soft mask of a given backbone model for new tasks. 
Comprehensive experiments on the ImageNet-to-Sketch dataset and twenty tasks of CIFAR-100 indicate that, with no need of using weight regularization and model expansion, the proposed method can run $\sim 10\times$ faster than the state-of-the-art CPG based learning method with similar accuracy performance. 
In addition, we analyze the effect of different backbone models. Even with a weak backbone model, the proposed method also could learn reasonable information for new tasks. We show that we can achieve better results compared with the related prior mask-based methods.
\clearpage
\footnotesize{
\bibliography{main_archiv.bib}
}

\end{document}